\pdfoutput=1

\documentclass[11pt]{article}

\usepackage[]{acl}

\usepackage{times}
\usepackage{latexsym}
\usepackage{verbatim}
\usepackage{fancyvrb}
\usepackage{rotating}
\usepackage{slashbox}
\usepackage{bm}

\usepackage[T1]{fontenc}

\usepackage[utf8]{inputenc}

\usepackage{microtype}

\usepackage{inconsolata}

\usepackage{graphicx}
\usepackage{booktabs}
\usepackage{colortbl}
\usepackage{multirow}
\usepackage{xspace}
\usepackage{amsmath}
\usepackage{xcolor}

\title{PixT3: Pixel-based Table-To-Text Generation}

\author{Iñigo Alonso, \\
    HiTZ Center - Ixa, \\
  University of the \\Basque Country UPV/EHU \\
  \texttt{inigoborja.alonso@ehu.eus} \\\And 
  Eneko Agirre \\
  HiTZ Center - Ixa, \\
  University of the \\Basque Country UPV/EHU \\
  \texttt{e.agirre@ehu.eus} \\\And
  Mirella Lapata \\
  Institute for Language, \\Cognition and Computation, \\
University of Edinburgh \\ 
  \texttt{mlap@inf.ed.ac.uk} \\}

\begin{document}
\maketitle
\begin{abstract}



  Table-to-text generation involves generating appropriate textual
  descriptions given structured tabular data. It has attracted
  increasing attention in recent years thanks to the popularity of
  neural network models and the availability of large-scale
  datasets. A common feature across existing methods is their
  treatment of the input as a string, i.e.,~by employing linearization
  techniques that do not always preserve information in the table, are
  verbose, and lack space efficiency.  We propose to rethink
  data-to-text generation as a visual recognition task, removing the
  need for rendering the input in a string format.  We present PixT3,
  a multimodal table-to-text model that overcomes the challenges of
  linearization and input size limitations encountered by existing
  models. PixT3 is trained with a new self-supervised learning
  objective to reinforce table structure awareness and is applicable
  to open-ended \emph{and} controlled generation settings. Experiments
  on the ToTTo \cite{parikh-etal-2020-totto} and Logic2Text
  \cite{chen-etal-2020-logic2text} benchmarks show that PixT3 is
  competitive and, in some settings, superior to generators that operate
  solely on text.\footnote{Our code, models, and data are available at \url{https://github.com/alonsoapp/PixT3}.}


\end{abstract}
    
\section{Introduction}
\label{sec:intro}
\begin{figure*}[t]
  \small
  \textbf{Table Title:} Shuttle America\\
    \textbf{Section Title:} Fleet\\
\includegraphics[scale=.235]{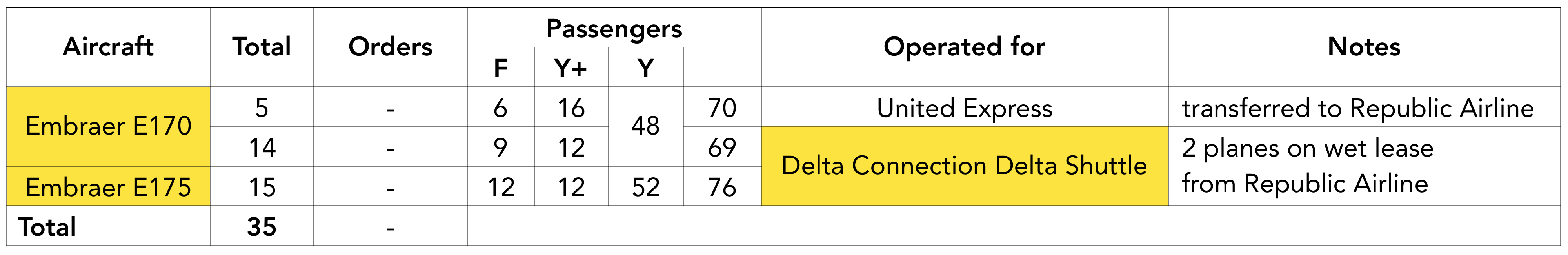}
  \vspace*{-.2ex}
\textbf{Linearized Table:}  \textsf{\small{<page\_title>
      Shuttle America <\/page\_title> <section\_title> Fleet
      <\/section\_title> <table> <row> <cell> Aircraft <\/cell> <cell>
      Total <row\_header> Aircraft <\/row\_header> <\/cell> <cell>
      Orders <row\_header> Aircraft <\/row\_header> <row\_header>
      Total <\/row\_header> <\/cell> <cell> Passengers <row\_header>
      Aircraft <\/row\_header> <row\_header> Total <\/row\_header>
      <row\_header> Orders <\/row\_header> <\/cell> <cell> Operated
      For <row\_header> Aircraft <\/row\_header> <row\_header> Total
      <\/row\_header> <row\_header> Orders <\/row\_header>
      <row\_header> Passengers <\/row\_header> <\/cell> <cell> Notes
      <row\_header> Aircraft <\/row\_header>}} $\dots \dots$ \\
\\ 
\raisebox{.3em}[0pt]{\textbf{Target
        Description:} Shuttle America operated the E-170 and the
      larger E-175 aircraft for Delta Air Lines.}
\caption{\label{tab:example} Example of table-to-text generation taken
  from the ToTTo dataset \cite{parikh-etal-2020-totto}. In the
  controlled setting, a natural language description is generated only
  for highlighted (yellow) cells. The table is linearized by encoding
  each value as a \textsf{(\small Column, Row, Value)} tuple. We only
  show the first row, for the sake of brevity.}
  \end{figure*}

Generating text from structured inputs such as tables, tuples, or
graphs, is commonly referred to as data-to-text generation
\citep{Reiter1997, Covington2001, Gatt2018}. This umbrella term
includes several tasks ranging from generating sport summaries based
on boxscore statistics \citep{Wiseman2017}, to producing fun facts
from superlative Wikipedia tables \cite{Korn2019}, and creating
textual descriptions given biographical data
\citep{lebret-etal-2016-neural}. From a modeling perspective,
data-to-text generation is challenging as it is not immediately
obvious how to best describe the given input. For instance, the table
in Figure~\ref{tab:example} can be verbalized in different ways,
depending on the specific content we choose to focus on.  In
\emph{controlled} data-to-text generation
\cite{parikh-etal-2020-totto}, models are expected to generate
descriptions for pre-selected parts of the input (see the
\emph{highlighted} cells in Figure~\ref{tab:example}).

Regardless of the generation setting, numerous approaches have emerged
in recent years with different characteristics. A few exploit the
structural information of the input \citep{puduppully-etal-2019-data,
  chen-etal-2020-kgpt, wang-etal-2022-robust}, use neural templates
\citep{Wiseman2018}, or resort to content planning
\citep{su-etal-2021-plan-generate, puduppully-2021-seq-plan}. While
others \citep{chen-etal-2020-logical, chen-etal-2020-logic2text,
  Aghajanyan2021,kasner-dusek-2022-neural} improve on fluency and
generalization by leveraging large-scale pre-trained language models
\citep{devlin-etal-2019-bert, 2020t5}. A common feature across these
methods is their treatment of tabular input as a string, following
various linearization methods.  As an example,
Figure~\ref{tab:example} shows the representation of tabular data
(top) as a sequence of (\textsf{\small Column, Row, Value}) tuples
(bottom).

Problematically, representing tabular information as a linear sequence results in a verbose representation that often exceeds the context window limit of popular Transformer models \cite{Vaswani2017}.
The challenge of processing such long sequences has fostered the
development of even more controlled methods which refrain from
encoding the table as a whole, concentrating exclusively on
highlighted content (e.g., \emph{only} the yellow cells in
Figure~\ref{tab:example}). Unfortunately, models trained on abridged
input have difficulty generalizing to new domains while being
practically ineffective in scenarios where content selection is not
provided.

 In this paper we propose to rethink data-to-text generation as a
 visual recognition task, allowing us to represent and preserve tabular information compactly. Vision Transformers (ViTs;
 \citealt{dosovitskiy2021an}) have significantly advanced the field of
 visual language understanding \cite{kim2022donut, davis2022end}
 demonstrating proficiency in various tasks, including language
 modeling \cite{rust-etal-2023-pixel}, visual document understanding
 \cite{Huang2022}, and visual question answering
 \cite{masry-etal-2022-chartqa}. Our work builds on Pix2Struct
 \cite{Lee2023}, a pretrained image-to-text model which can be
 fine-tuned for visually-situated language tasks. We recast
 data-to-text generation as an image-to-text problem and present
 PixT3, a \textbf{Pix}el-based
 \textbf{T}able-\textbf{t}o-\textbf{T}ext model, which is generally
 applicable to open-ended and controlled generation settings,
 overcoming the challenges of linearization and input size limitations
 encountered by existing models.

Our contributions can be summarized as follows:
(a)~we introduce the first pixel-based model for table-to-text
generation and showcase its robustness across generation settings with
varying table sizes; (b)~we propose a new training curriculum and
self-supervised learning objective to reinforce table structure
awareness; (c)~automatic and human evaluation results on the ToTTo benchmark \cite{parikh2020totto}
show that PixT3 excels in open-ended generation, leading to improved
faithfulness and generation quality, while being competitive with
existing methods in controlled scenarios;
and (d)~we present a new dataset based on Logic2Text
\cite{chen-etal-2020-logic2text}, which allows us to evaluate
generalization capabilities of current table-to-text models.

\section{Related Work}
\label{sec:related-work}

The bulk of previous work treats tables as textual objects. Several
techniques have been developed to extract accurate information from
them \citep{puduppully-etal-2019-data, chen-etal-2020-kgpt} using
templates \citep{Wiseman2018}, enforcing table structure awareness
\cite{mahapatra-garain-2021-exploring, wang-etal-2022-robust},
applying contrastive learning \cite{An2022, chen-etal-2023-table} or
  focusing on content planning \citep{su-etal-2021-plan-generate,
    puduppully-2021-seq-plan}.  Other techniques
  \citep{chen-etal-2020-logical, chen-etal-2020-logic2text,
    Aghajanyan2021,kasner-dusek-2022-neural} improve fluency and
  generalization by leveraging large-scale pretrained language models
  \citep{devlin-etal-2019-bert, 2020t5}.
  Tables are generally linearized, even when special-purpose
  techniques are developed for encoding table structure
  \cite{wang-etal-2022-robust}. Dedicated table understanding
  techniques \cite{tuta-wang-2021,jin-etal-2023-tabprompt} eschew
  linearization but have not been integrated with generation tasks.

  Previous attempts to address table-to-text generation from a visual
  recognition perspective \cite{dash2023,Srihari2003} have relied on
  OCR methods which first extract text from the image and then feed it
  as a string to a generation model. Aside from being noisy, these
  techniques typically embrace a text-centric point of view, treating
  the image as a limitation rather than an informative modality. Our
  work builds on recent visual language understanding models
  \cite{kim2022donut,davis2022end,Lee2023} which are based exclusively
  on pixels and have managed to outperform OCR methods in several
  natural language processing tasks
  \cite{rust-etal-2023-pixel,Huang2022,masry-etal-2022-chartqa,salesky-etal-2023-multilingual}.

The field of Vision Language Models (VLMs) has also experienced significant growth in recent years \cite{liu2023visual, ye2023mplugowl2, Qwen-VL, wang2023cogvlm, alayrac2022flamingo}. While most of them focus primarily on natural images, a few are starting to explore the application of dual encoder architectures to visually represented language \cite{ye-etal-2023-ureader, zhang2023llavar}. However, these architectures are not parameter lean 
(with increased model size of a factor of~40 or more compared to Pix2Struct), and some continue to rely on fixed resolution images which can be particularly problematic when processing tabular data.

A few other efforts have recently explored multimodal approaches to
processing tables for various tasks, including table-to-text
generation.  \citet{dash2023} convert images into HTML tokens which
are subsequently linearized and processed by a traditional
text-to-text model. Other work \cite{chen-etal-2023-tablevlm} focuses
on recognizing the structure of tables from images as an independent
task.  It also leverages multimodal pretraining and unsupervised table
structure learning objectives, but ignores the content of table cells
and their relations. To the best of our knowledge, our work is the
first to conceptualize data-to-text generation as a visually-situated
language understanding problem.





\section{Problem Formulation}
\label{sec:formulation}

The task of table-to-text generation aims to take a structured
table~$\bm{t}$ as input and output a natural language description
$\bm{y} = [y_1, \dots, y_k]$ where~$k$ is the length of the
description. Table~$\bm{t}$ is typically reformatted as a sequence
of textual records \mbox{$\bm{t} = [t_{1,1}, t_{1,2}, \dots, t_{i,j},
    \dots, t_{m,n}]$} where~$m$ and $n$~respectively denote the number
of rows and columns of~$\bm{t}$.

We approach this task from a visual recognition perspective, and expect
the input table to be an image~$\bm{x}$.
 The image is reshaped into a sequence of patches analogous
to linguistic tokens. More formally, for an input image $\bm{x} \in
R^{H\times W \times C}$ and patch size~$p$, we create~$N$ image
patches denoted as $x_p \in R^{N \times (P^{2} \cdot C)}$.  $(H,W)$ is
the resolution of the original image, $C$ is the number of channels,
$(P,P)$ is the resolution of each image patch, and $N = \frac{HW}{P^2}$
the resulting number of patches, which serves effectively as the input
sequence length. Our proposed model learns to autoregressively
estimate the conditional probability of a text sequence from a source
image as:
\begin{equation}
  \label{equation:1}
P(\bm{y}|\bm{x};\bm{\theta}) = \prod\limits_{i=1}^{n} P(y_i|\bm{y}_{<i},\bm{x};\bm{\theta})
\end{equation}
where~${\bm \theta}$ are transformer parameters and $\bm{y}_{<i}$~the
  words decoded thus far. 

We further define three generation settings, which manipulate the
information provided to the model in terms of content selection (see
Appendix~\ref{sec:table-text-gener} for visualization). In the
\emph{tightly-controlled} setting (TControl), the model is given
highlighted cells only, ignoring the table. Most recent approaches
benchmark model performance in this setting
\cite{wang-etal-2022-robust,An2022,
  chen-etal-2023-table,su-etal-2021-plan-generate,kale-rastogi-2020-text}. In
the \emph{loosely controlled} setting (LControl), the model is given
highlighted cells \emph{and} the entire table. This is the original
setting for which the ToTTo dataset \cite{parikh-etal-2020-totto} was
constructed. Finally, we introduce the \emph{open-ended} setting
(OpenE), where the model is given the table without any highlighting.



\begin{figure}[t]
\centering
\includegraphics[scale=.4]{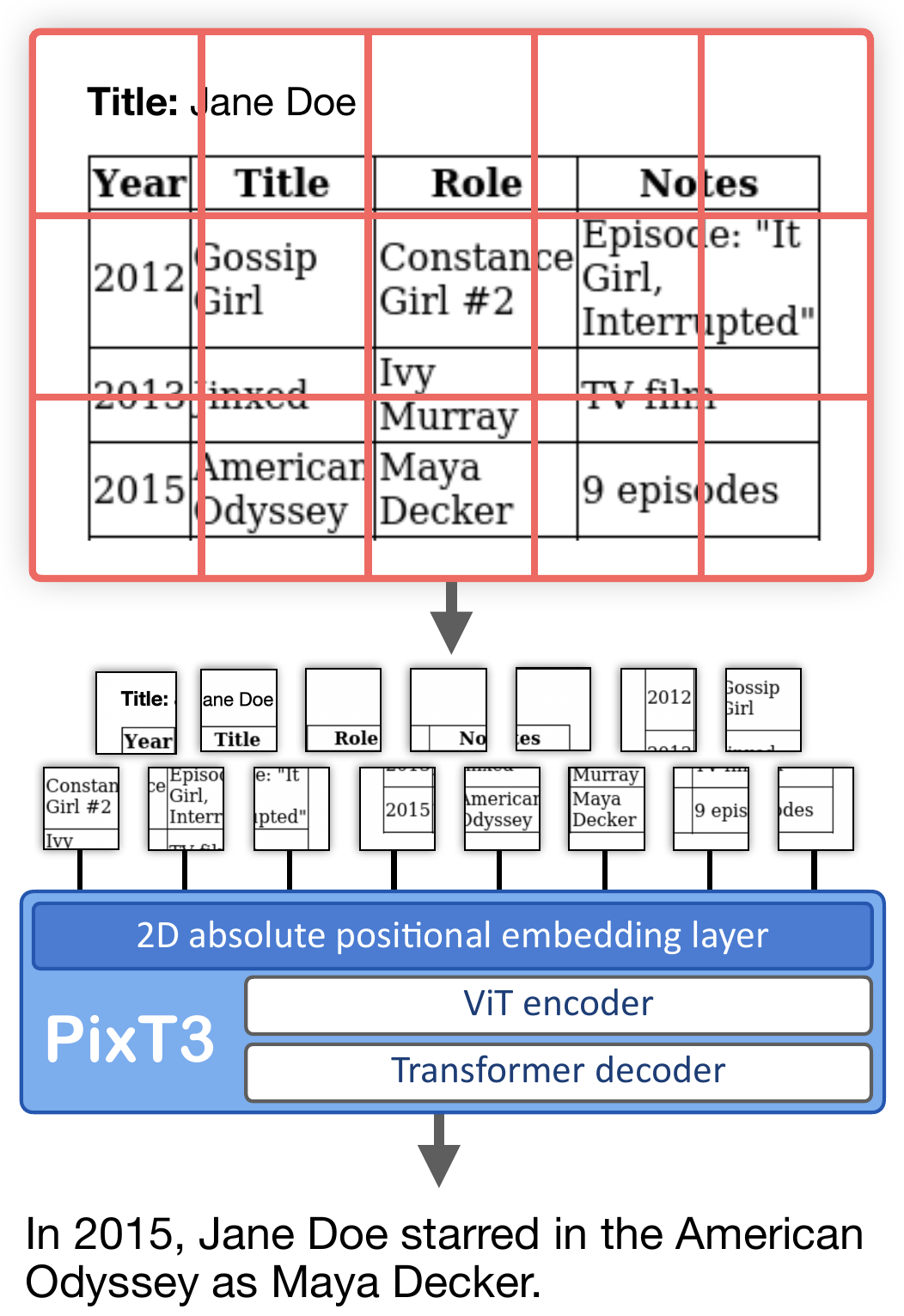} 
\caption{Overview of PixT3  generation model.}
\label{fig:pixt3_overview}
\end{figure}

\section{The PixT3 Model}

PixT3 is an image-encoder-text-decoder model based on Pix2Struct
\cite{Lee2023}. It expects image rendered tables and generates
descriptions thereof (see Figure~\ref{fig:pixt3_overview}).
Pix2Struct is a Vision Transformer model pretrained on 80M~screenshots
of web pages extracted from URLs in the C4 corpus \cite{2020t5}. It
splits input images into patches of \mbox{16$\times$16} pixels (see
Figure~\ref{fig:pixt3_overview}), linearly embeds each patch, adds
position embeddings, and feeds the resulting sequence of vectors to a
standard Transformer encoder \cite{Vaswani2017}.

Pix2Struct was first warmed up with a reading curriculum
\cite{rust-etal-2023-pixel,davis2022end}, to improve training
stability and fine-tuning performance and then pretrained with a
screenshot parsing objective; specifically, it generates a simplified
version of an HTML subtree that represents a highlighted area of a web
page screenshot.  It also adds a \mbox{BART-lik}e
\cite{lewis-etal-2020-bart} learning signal to pretraining by
masking~50\% of the text in the input and then requiring the model to
produce the entire subtree.
Importantly for our table-to-text generation task, Pix2Struct supports
variable image resolution and multiple aspect ratios. It first
re-scales the input (up or down) to extract the maximal number of
fixed-size patches that fit within a given sequence length and then
replaces the typical \mbox{1-dimensional} absolute positional
embedding with a \mbox{2-dimensional one}, which adds resolution
flexibility and removes any aspect ratio distortion.





We initialize PixT3's model weights with Pix2Struct; we next adopt a
curriculum training strategy which instills in our model knowledge
about tables and their structure (see Section \ref{sec:warmup}); and
finally, we fine-tune on table-to-text generation datasets such as
ToTTo \cite{parikh-etal-2020-totto} with a task-specific supervised objective.

 
\subsection{Table-to-Image Rendering} 
\label{sec:rendering}



We parse tables to HTML, and subsequently render them into images. We
also render table metadata (e.g., Wikipedia page and section title),
if it exists, as part of the image, adding it on top of the
table. Tables are rendered into three different images corresponding
to the generation settings defined in Section~\ref{sec:formulation}
(see Appendix~\ref{sec:table-text-gener}, Figure~\ref{fig:examples}). 
%



Although Pix2Struct can handle variable resolutions and input patches,
very long inputs are nevertheless computationally expensive.
Following \citet{Lee2023}, we set the maximum input length to
2,048~patches (of~\mbox{16$\times$16} pixels) which corresponds to a
maximum image size of~524,288 pixels.  41.74\% of the tables in a
dataset like ToTTo \cite{parikh-etal-2020-totto} exceed this size (see
Figure \ref{fig:size_dist_log} in
Appendix~\ref{app:size_performance}), with 5\%~being larger than 8.3M
pixels (32,768 patches). Indiscriminately down-scaling \emph{all}
images exceeding the maximum input length would negatively affect
performance, especially for very big tables, effectively rendering
them unreadable (we showcase how image size affects model performance
in Figure~\ref{fig:size_performance}). To avoid this as much as
possible, we truncate the image to fit within a maximum down-scaling
factor~$\gamma$. In other words, images are first compressed to
$\gamma\%$~of their original size and then truncated from left to
right until they fit into 2,048 patches. The optimal value
for~$\gamma$ is determined empirically (see
Appendix~\ref{sec:scale_truncation}).




\subsection{Structure Learning Curriculum}
\label{sec:warmup}


Pix2Struct is a general-purpose visual language understanding model,
and as such it is not particularly knowledgeable about tables and
their structure. Tables can be presented in a variety of ways
visually, such as spanning multiple columns or rows, with or without
horizontal and vertical lines, non-standard spacing and alignment, and
text formatting. Aside from presentation, there are various
conventions about the underlying semantics of tables and their
structure, e.g.,~each cell is only related to cells in the same column
and row. These challenges have led to the development of dedicated
table understanding techniques \cite{jin-etal-2023-tabprompt,
  wang-etal-2022-robust} in the domain of text but cannot be readily
ported to images.

Instead, we encourage PixT3 to adhere to tabular conventions, by first
training it on an intermediate supporting task.  This acts as a
structure learning curriculum, exposing the model to the rules
governing tables. We next elaborate on the intermediate task, its
corresponding dataset, and the proposed self-supervised
objective. 



\paragraph{Dataset for Intermediate Training}

Existing datasets like \mbox{ICDAR2021} \cite{kayal2021icdar} and
TableBank \cite{li2019tablebank} are representative of the task of
parsing table images into their structure and, in theory, could be
used for our intermediate training purposes.  However, they focus on
scientific tables which do not follow the typical distribution of
Wikipedia tables found in ToTTo \cite{parikh-etal-2020-totto},
e.g.,~in terms of size and cells spanning across rows and columns.
We instead propose to create a synthetic image-to-text dataset,
making use of the table rendering pipeline described in
Section~\ref{sec:rendering}. Although we generate tables specifically
tailored for our use-case, the generation process is flexible and can
be adapted to other domains with different characteristics.

\begin{figure}[t]
\includegraphics[scale=.25]{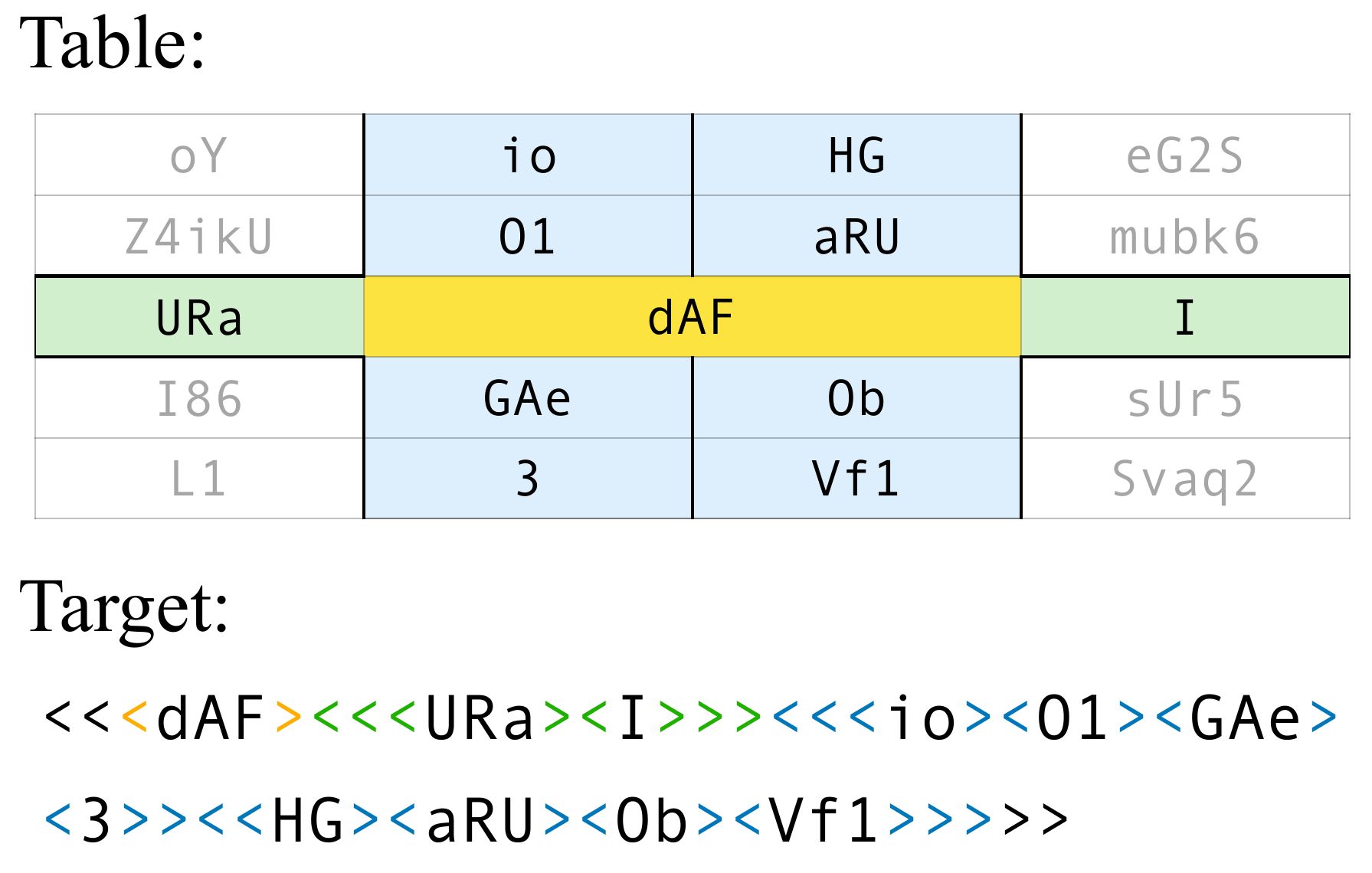}
\caption{\label{fig:synthetic} 
Synthetically generated table with a highlighted cell and corresponding pseudo-HTML  target sequence (for self-supervised  objective). Cells within the target sequence are highlighted in the table with a colored background. For details on the structure of the target, please refer to Appendix \ref{app:ssl3}.}

\end{figure}

We determine the structure of each table (size, column, and row spans)
randomly, following ToTTo's training set distribution. We cap the
generation process at a maximum of 20~columns and~75 rows. Table cells
are filled with synthetic values consisting of a random combination of
one to five random English alphabet characters and digits, functioning
as identifiers rather than meaningful values (see
Figure~\ref{fig:synthetic} for an example).  Our 
dataset contains 135,400~synthetic tables, 120,000 for training, 7,700
for validation, and 7,700 for testing.

\paragraph{Self-supervised Objective}
\label{sec:ssl1}

While masking is a widely adopted learning objective
\cite{devlin-etal-2019-bert}, it does not naturally transfer to our
table-to-text generation task; table values are not naturally
correlated to neighboring values and thus a masked cell cannot be
easily predicted from other cells in its context. Table values could
be rearranged so that they correlate to their neighbors, however,
early experiments showed that this type of objective does not improve
downstream task performance (see Appendix~\ref{app:ssl3} for
details). Another common pretraining objective is table linearization
\cite{chen-etal-2023-tablevlm}, which, however, scales poorly with
table size, leading to slow pretraining.


We propose a self-supervised objective that encourages PixT3 to
capture the relations between cells within a table while generating a
small amount of tokens. Specifically, we highlight a random cell in a
synthetically generated table, and train the model to produce a sorted
list of cells within the same column and row (see
Figure~\ref{fig:synthetic}). Our objective encapsulates a loose notion
of table structure, nudging the model to pay attention to the
arrangement of columns and rows around a cell.  We follow the same
pseudo HTML notation introduced in Pix2Struct to format our output
sequence, easing the model's transition from its original screenshot
parsing objective to this new one. Note that we consider tables with a
heterogeneous structure where cells can span across multiple columns
and rows. In such cases, the expected sequence will contain all cells
in related rows and columns surrounding the highlighted cell (see
Figure~\ref{fig:synthetic}).




\subsection{PixT3 Fine-tuning}
\label{sec:fine-tuning}
The intermediately pre-trained PixT3 is subsequently fine-tuned on an image-rendered
dataset (see Section~\ref{sec:rendering}). In experiments, we use ToTTo \cite{parikh2020totto}, however, our approach is not tied to a particular style of tables. Due to our model's requirement for unimodal input, we treat table-related information (such as its title) as part of the table itself and render them both as one image (see  \citealt{Lee2023} for a similar approach). 

\section{Experimental Setup}
\label{sec:experimental-setup}

\paragraph{Model Configuration}
All our experiments were conducted with the \emph{base} pretrained
Pix2Struct\footnote{\footnotesize{\mbox{\url{https://github.com/google-research/pix2struct}}}}
model (282M~parameters).  We trained PixT3 variants for the three
table-to-text generation settings defined in
Section~\ref{sec:formulation}. All PixT3 models were fine-tuned on
ToTTo \cite{parikh-etal-2020-totto} with tables rendered as images
following the procedure outlined in Section~\ref{sec:rendering}. The
maximum down-scaling factor~$\gamma$ was set to 0.39.

PixT3 models were fine-tuned with a batch size of~8 and a gradient
accumulation of~32 steps on a single NVIDIA A100 80GB~GPU. Checkpoints were
selected according to best performance on the validation set. All
models used an input sequence length of 2,048~patches and were
optimized with AdamW \cite{loshchilov2017decoupled}. We used a
learning rate scheduler with a linear warmup of~1,000 steps to~0.0001,
followed by cosine decay to~0. The decoder maximum sequence length was
set to~50 tokens, which covers 97.49\%~of the target descriptions in
the training data. PixT3 was trained for 1.4k~steps with the
self-supervised objective described in Section~\ref{sec:warmup}. Our
decoder was not frozen during intermediate training, as initial
experiments showed that a fully trained model outperformed one with
frozen decoder weights. A full list of fine-tuning hyper-parameters can be found in Appendix~\ref{app:hyperparams}.

\paragraph{Datasets}
We evaluated our model on ToTTo \cite{parikh-etal-2020-totto}, a
large-scale, manually curated dataset representative of several
domains and types of tables.  We also assessed the generalization
capabilities of PixT3 on out-of-distribution tables. We created an
out-of-domain benchmark with content selection annotations similar to
ToTTo based on Logic2Text \cite{chen-etal-2020-logic2text}, an
existing dataset which contains a total of~10,161 Wikipedia
tables, paired with human-authored descriptions and logical forms.
Logic2Text differs from ToTTo in that descriptions are not simple
verbalisations of table rows and columns, but require some form of
reasoning (e.g., comparisons or counting operations).  We were able to
automatically trace values mentioned in the logical form back to the
cells of the input tables \cite{alonso2023automatic}, thus obtaining
highlighted cell annotations similar ToTTo's (see
Appendix~\ref{sec:addit-results-exampl} for an example). We report
results on the official test set (1,085 examples).

%

\paragraph{Model Comparison} 
We evaluated PixT3 against several text-only models with similar
parameter sizes. These include CoNT \cite{An2022}, the top performant
(published) model in the ToTTo leaderboard.\footnote{A model named SKY
  appears to slightly outperform CoNT in the leaderboard, however, at
  the time of writing, we were not able to verify this, i.e.,~by
  finding a publication or preprint describing this model.} CoNT is a
text-to-text generation model which makes use of contrastive learning,
through improved selection of contrastive examples, a new contrastive
loss, and a global decoding strategy. CoNT expects the input table to
be converted to a string, and is built on top of \mbox{T5-base} (220M
parameters). We also compared against Lattice
\cite{wang-etal-2022-robust}, a model which enforces awareness of
table layout though pruning the attention flow and encoding cells in a
way that is invariant to their relative position in a sequence. This
model also uses \mbox{T5-base} and expects linearized input.  In
addition, we report results with vanilla \mbox{T5-base} which
performed competitively on the ToTTo leaderboard without any task
specific modifications \cite{kale-rastogi-2020-text,An2022}.  All
comparison models and PixT3, were trained on the ToTTo training set in
our three generation settings.\footnote{Comparison models were
  trained with the authors' publicly available scripts.}



For our out-of-domain experiments,  we also compare against LLaVA-1.5
\cite{liu2023visual}, a large pretrained multimodal model (13B
parameters) which is built on top of the CLIP visual encoder
\cite{DBLP:conf/icml/RadfordKHRGASAM21} and the 
Vicuna-7B language model \cite{zheng2023judging}, 
and fine-tuned on vision-language
instructions.  LLaVA has not been fine-tuned specifically for
table-to-text generation, however, it is interesting to see if
sufficiently large scale is all it takes to do well on the table-to-text
generation task. 
LLaVA can only handle a \emph{single} image at each forward pass. This limitation prevents it from performing inference in an in-context learning setting, where  the model has access to multiple input-output examples at the same time. To approximate in-context learning as closely as possible, we provided LLaVA with an image, an instruction, and three table descriptions as output examples for each generation setting (see Appendix~\ref{sec:llava_promts} for details). We summarize the number of parameters for all comparison models in Table~\ref{tab:logic2text_results}.

we do still provide a few description examples in our prompt to ensure a fair zero-shot comparison. All prompts used for LLaVA in this evaluation can be found in Appendix F.



\begin{table}[t]
\centering
\begin{small}
\begin{tabular}{@{}l@{~}l@{~~~}c@{~~~}c@{~~~}c@{~~~}c@{~~~}c@{~~~}c@{~~~}c@{~~~}c@{~~~}c@{}}
\toprule
& \textbf{}                   & \multicolumn{2}{c}{{Dev}} & \multicolumn{2}{c}{{TestN}} & \multicolumn{2}{c}{{TestO}} & \multicolumn{2}{c}{{Test}} \\ \midrule
& {Model}              & {BL}        & {PR}        & {BL}        & {PR}         & {BL}         & {PR}         & {BL}        & {PR}         \\ \midrule
& T5-base                     & 47.7                 & 57.1                & 38.9                 & 51.2                 & 55.4                  & 61.1                 & 47.2                 & 56.2                 \\
& T5-3B                       & 48.4                 & 57.8                & 39.3                 & 51.6                 & 55.1                  & 60.7                 & 47.2                 & 56.2                \\
& Lattice                     & 48.0                 & 58.4                & 40.0                  & \textbf{53.8}                  & 55.9                   & 62.4                  & 48.0                  & \textbf{58.1}                \\
& CoNT                        & \textbf{49.0}       & \textbf{58.6}      & \textbf{40.6}                 & 53.7                 & \textbf{56.7}                  & \textbf{62.5}                 & \textbf{48.7}                & \textbf{58.1}       \\
\raisebox{.8em}[0pt]{\begin{sideways}TControl\end{sideways}} & PixT3                       & 45.7                 & 55.7                & 37.5                    & 50.6                    & 53.2                     & 60.4                    & 45.4                    & 55.5                    \\ \midrule
& T5-base                     & 24.5                 & 27.2                & 19.4                 & 23.9                 & 29.4                  & 30.3                 & 24.5                 & 27.1                 \\
& T5-3B                       & 23.6                 & 26.0                & 18.0                 & 22.4                 & 28.7                  & 29.2                 & 23.4                 & 25.8                 \\
& Lattice                     & 24.9                 & 31.0                & 20.8                  & 27.7                  & 27.5                   & 33.8                  & 24.4                  & 30.8                \\
& CoNT                        & 23.8                 & 29.3                & 19.2                 & 26.1                 & 28.7                  & 32.3                 & 23.9                 & 29.2                 \\
\raisebox{.8em}[0pt]{\begin{sideways}LControl\end{sideways}} & PixT3  & \textbf{46.2}        & \textbf{55.1}       & \textbf{38.1}        & \textbf{50.3}        & \textbf{52.7}         & \textbf{59.0}                 & \textbf{45.4}        & \textbf{54.7}        \\ \midrule
& T5-base & 21.5                 & 23.5                & 16.8                 & 21.0                 & 26.5                  & 26.5                 & 21.7                 & 23.8                 \\
& T5-3B   & 20.8                 & 22.9                & 16.7                 & 20.3                 & 25.5                  & 25.5                 & 21.2                 & 22.9                 \\
& Lattice & 20.9                 & 26.1                & 17.6                  & 24.3                  & 23.7                   & 27.6                  & 20.8                  & 25.9                \\
& CoNT    & 21.7                 & 25.8                & 16.9                 & 23.2                 & 26.3                  & 28.3                 & 21.6                 & 25.8                 \\
\raisebox{1em}[0pt]{\begin{sideways}OpenE\end{sideways}} & PixT3   & \textbf{24.8}        & \textbf{28.3}       & \textbf{20.5}        & \textbf{26.3}        & \textbf{28.9}         & \textbf{30.3}        & \textbf{24.7}        & \textbf{28.3}       \\ \bottomrule
\end{tabular}
\end{small}
    \caption{\label{tab:totto_results} Automatic evaluation results on
      ToTTo in three generation settings: tightly controlled
      (TControl), loosely controlled (LControl), and open-ended
      (OpenE). 
      We report BLEU (BL) and PARENT (PR) results
      on the development (Dev) and Test sets, including the
      overlapping (TestO) and non-overlapping (TestN) test set
      splits. BLEURT results are in
      Appendix~\ref{sec:addit-results-exampl}.}
\end{table}

\section{Results}
\label{sec:results}

\paragraph{PixT3 is the best performing model in loosely controlled and open-ended
  generation settings.}


Table~\ref{tab:totto_results} summarizes our results on ToTTo in our
three generation settings. We evaluated model performance
automatically with the same metrics used to rank participant systems
in the ToTTo leaderboard. These include BLEU
\cite{papineni-etal-2002-bleu} which is as a proxy for fluency, PARENT
\cite{dhingra-etal-2019-handling}, a metric proposed specifically for
data-to-text evaluation that takes the table into account, serving as
a proxy of faithfulness, and BLEURT \cite{sellam-etal-2020-bleurt};
the latter is a composite metric that takes a reference and model
output as input, and returns a score that indicates the extent to
which the output is fluent and conveys the meaning of the reference.
Note that ToTTo features two splits in the development/test set
containing tables whose header values are present (overlapping split)
and absent (non-overlapping split) in the training set. Results on the
test set, which is not publicly available, were obtained via
submitting to the ToTTo leaderboard.




We first discuss our results on the tightly controlled generation
setting (TControl) where models are not given the full table, just the
highlighted cells. We would not expect PixT3 to excel at this setting,
which is better suited to text-to-text models (highlighted cells make
for non-descriptive images, see Appendix~\ref{sec:table-text-gener},
Figure~\ref{fig:examples}). PixT3 is indeed unable to outperform CoNT,
Lattice, and related T5 variants, falling 3.5 BLEU points behind on
the development set and 3.7 on the test set. However, LControl, the
loosely controlled generation setting, better showcases the advantages
of PixT3, which in this case demonstrates almost a two times
improvement over CoNT and T5 models.  Performance degrades drastically
for all systems in the open-ended setting (OpenE) which is
challenging; models are expected to perform content selection in
addition to text generation, and could produce table descriptions
which are valid but different from the reference.  Automatic metrics
based on n-gram overlap are particularly punitive in this
case. Nevertheless, PixT3 is superior to CoNT, Lattice, and T5 across
evaluation metrics.



\begin{table}[t]
\begin{small}
\begin{center}
  \begin{tabular}{@{}llrrc@{}} \toprule
& {Model} & {Size} & \multicolumn{1}{c}{BLEU} & {PARENT} \\ \midrule
& LLaVA          & 13B          & 12.6          & 34.36          \\
& T5-base        & 220M          & 16.8          & 55.97          \\
& T5-3B          & 3B          & 17.7          & 52.75          \\
& Lattice        & 220M          & 19.8          & 61.05          \\
& CoNT           & 220M          & 18.8          & 61.73          \\
\raisebox{1.2em}[0pt]{\begin{sideways}TControl\end{sideways}} & PixT3          & 282M          & \textbf{20.6} & \textbf{61.86}  \\ \midrule
& LLaVA          & 13B          & 5.9           & 23.18          \\
& T5-base        & 220M          & 11.5          & 40.02         \\
& T5-3B          & 3B          & 10.9          & 35.45          \\
& Lattice        & 220M          & 11.5          & 40.02          \\
& CoNT           & 220M          & 11.8          & 43.25          \\
\raisebox{1.2em}[0pt]{\begin{sideways}LControl\end{sideways}} & PixT3          & 282M          & \textbf{21.5} & \textbf{56.45}  \\ \midrule
& LLaVA          & 13B          & 6.7           & 20.14          \\
& T5-base        & 220M          & 7.9           & 30.67          \\
& T5-3B          & 3B          & 9.5           & 29.47          \\
& Lattice        & 220M          & \textbf{11.7} & \textbf{38.12} \\
& CoNT           & 220M          & 11.0          & {36.94}        \\
\raisebox{1.6em}[0pt]{\begin{sideways}OpenE\end{sideways}} & PixT3          & 282M          & {11.4} & 35.68          \\ \bottomrule
  \end{tabular}
\end{center}
\end{small}
    \caption{Automatic evaluation results on Logic2Text in three
      generation settings: tightly controlled (LControl), loosely
      controlled (LControl), and open-ended (OpenE). All models
      (except LLAVA) were fine-tuned on ToTTo
       and tested on Logic2Text. BLEURT results are in Appendix~\ref{sec:addit-results-exampl}.}
    \label{tab:logic2text_results}
\end{table}

\paragraph{PixT3 generalizes to out-of-domain tables which require
  reasoning skills.}

We next evaluate whether PixT3 generalizes to unseen tables, outside
ToTTo's distribution.  Table~\ref{tab:logic2text_results} shows our
results on Logic2Text \cite{chen-etal-2020-logic2text}, again
following the three generation settings.  Compared to ToTTo,
Logic2Text is a more challenging dataset as most descriptions rely
on reasoning over the entire table. This results in poor model
performance in the TControl setting which does not include the table
as input. Nonetheless, we observe that PixT3 excels at the
LControl setting, even though it has to process and reason over the
entire table. The OpenE setting is challenging for all models as they
are asked to identify interesting cells to talk about in
\emph{out-of-domain} tables. PixT3 still maintains an edge over T5 and
LLaVA, performing on par with CoNT and Lattice. We observe that LLaVA
cannot match the performance of PixT3 and T5-based models. This
underscores the importance of task-specific fine-tuning over parameter
size.  We present output examples in
Appendix~\ref{sec:addit-results-exampl}.

 \begin{figure}[t]
   \vspace*{-.7cm}
\includegraphics[width=\linewidth]{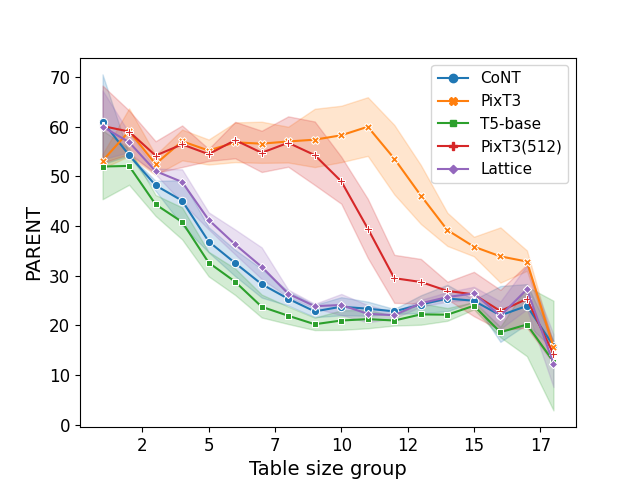}
\caption{Model performance (CoNT, T5, PixT3, Lattice, and PixT3 with
  512 patch input size) in the LControl setting across 18 table size
  groups (logarithmic scale). Upper and lower bounds in  shaded
  areas correspond to results for the overlapping and non-overlapping
  ToTTo splits, while central points correspond to results overall. We
  report results with PARENT, other metrics show similar tendencies.
  We refer to Appendix~\ref{app:size_performance} for further
  details.}
\label{fig:size_performance}
\end{figure}

\paragraph{PixT3 is robust against table input size.} In 
Figure~\ref{fig:size_performance}, we analyze the effect of table size
on model performance. As can be seen, T5, Lattice, and CoNT are severely
affected: the bigger the table, the less accurate the generated
description. PixT3 is evidently more robust, showing degradation in
performance only for very big tables. We also examined whether PixT3
has an edge because of its ability to encode longer inputs. Recall
that CoNT, Lattice, and T5-base utilize a fixed input length of~512
tokens, while PixT3 uses~2,048 patches. We thus trained a PixT3
variant with input length set to~512 patches. As shown in
Figure~\ref{fig:size_performance}, the more constrained PixT3 model is
slightly worse and more likely to degrade with increased table size
but consistently outperforms CoNT, Lattice, and T5.

\paragraph{The structure learning curriculum improves generation
  quality across metrics.}  In Table~\ref{tab:ablation} we perform an
ablation study comparing PixT3 with and without our structure learning
curriculum and self-supervised objective (Section~\ref{sec:warmup}).
For both models we follow the same fine-tuning process: we render
tables into images, identify the optimal point of image compression
and truncation (see Section~\ref{sec:rendering}), and perform
hyper-parameter search to optimize Pix2Struct-base for our task.
Vanilla PixT3 (second row in Table~\ref{tab:ablation}) shows a
substantial improvement over an out-of-the-box Pix2Struct model which
achieves a BLEU score of~0.2 and PARENT score of~0.6 on the ToTTo
development set.  Adding the intermediate training curriculum (second
row in Table~\ref{tab:ablation}) slightly improves vanilla PixT3
across evaluation metrics. 

Manual inspection of the descriptions produced by
the two PixT3 model variants reveals they are often
semantically equivalent to the target (43\% of the time). 
Nevertheless, the intermediate training curriculum substantially reduces structure-based faithfulness errors, especially in the OpenE setting. On a sample of 200~outputs (randomly selected from the development set), we found that 23\%~of the descriptions produced by vanilla PixT3 disregard or misinterpret the structure of the table.  Structural faithfulness errors reduce to~7\% when PixT3  is trained with our structure learning curriculum.


\begin{table}[t]
\centering
\begin{small}
\begin{tabular}{@{}lr@{~~}r@{~~}r@{~~}c@{~~}c@{~~}c@{}}
\toprule
                     & \multicolumn{3}{c}{{Dev}} & \multicolumn{3}{c}{{Test}} \\\midrule
 \multicolumn{1}{c}{Models}              & {BL}        & {PR} &  BRT  &     {BL}        & PR         & {BRT}        \\ \midrule
Pix2Struct & {}{}0.2 & 0.6 & $-$1.433   & ---  &--- & ---\\
PixT3 (W/o SLC) & 38.7 & 46.0 & $-$0.003   & 38.3  &45.6 & 0.001\\
PixT3 (With SLC)      & {\bf 39.2} & {\bf 46.5} &  {\bf 0.008} &  {\bf 38.7}  &
{\bf 46.3}& {\bf 0.007} \\ \bottomrule
\end{tabular}
\end{small}
\caption{\label{tab:ablation} PixT3 with and without structure
  learning curriculum (SLC); we report results on the ToTTo
  development (Dev) and Test set with BLEU (BL), PARENT (PR), and
  BLEURT (BRT), averaged across the three generation settings.} 
\end{table}

\paragraph{PixT3 is most faithful
  in loosely controlled and open-ended generation settings.}  We
further conducted a human evaluation study to quantify the extent to
which the generated descriptions are faithful to the table.  We
evaluated PixT3, and the two best performing text-only systems
(CoNT, and Lattice) on two sets of 100~randomly selected
table-description pairs from ToTTo (development set) and Logic2Text
(test set), in the three generation settings.  Crowdworkers were
presented with an uncompressed image of a table, its page and section
title, and a model generated description. As an upper bound, we also
elicited judgments for the human curated reference descriptions for
the same ToTTo and Logic2Text examples. Participants were asked to
determine whether a description was "True" or "False" based on the
information provided in
the table and/or its title and subtitle (see instructions in Appendix \ref{app:form}). Overall we elicited~7,200
judgments (100 examples $\times$ 3 generation settings $\times$ 4
model descriptions $\times$ 3 participants $\times$ 2 datasets).
Crowdworkers were recruited using the online platform
Prolific.\footnote{\url{https://www.prolific.com}}

Table~\ref{tab:human_eval} shows the results of the human evaluation,
specifically the proportion of descriptions deemed faithful.  As
expected, the human authored Reference description is consistently
faithful across generation settings. CoNT is more faithful in TControl
but deteriorates in the LControl and OpenE settings. We further
examined whether differences among systems are statistically
significant using paired bootstrap resampling. PixT3 is significantly
worse ($p<0.05$) than the Reference in TControl but not CoNT or Lattice. In LControl all differences between systems are statistically significant
($p<0.05$). In OpenE, PixT3 is significantly different ($p<0.05$) from
CoNT and Lattice but not from the Reference. Inter-rater agreement was moderate
with a Fleiss' Kappa coefficient of~0.55 \citep{fleiss1971measuring}.


\begin{table}[t]
\centering
\begin{small}
\begin{tabular}{@{}llccc@{}}
\toprule
& {Model} & {TControl} & \multicolumn{1}{c}{LControl} & {OpenE} \\ \midrule
& Reference      & 87               & 84              & 89           \\
& Lattice & {\bf 79} & 16 & 20 \\
& CoNT           & 76      & 16              & 35           \\
\raisebox{.6em}[0pt]{\begin{sideways}ToTTo\end{sideways}} & PixT3    &
         {69}                & {\bf 72}     & {\bf 78}  \\ \midrule
& Reference      & 81               & 87              & 86           \\
& Lattice &  34 & {}{~~}3& 16\\
         & CoNT           & {\bf 35}      & {}{~~}3              & 26           \\
\raisebox{1em}[0pt]{\begin{sideways}L2T\end{sideways}} & PixT3    &
         {32}                & {\bf 40}     & {\bf 60}  \\ \bottomrule
\end{tabular}
\end{small}
    \caption{Human evaluation results on ToTTo and Logic2Text
      (L2T). Proportion of descriptions rated as faithful for PixT3,
      CoNT, and Reference in three generation settings: tightly
      controlled (LControl), loosely controlled (LControl), and
      open-ended (OpenE).}
    \label{tab:human_eval}
\end{table}


\section{Conclusion}
\label{sec:conclusion}

In this paper, we leverage the capabilities of Vision Transformers to
recast table-to-text generation as a visual recognition task, removing
the need for rendering the input in a string format.  Our model,
PixT3, introduces a new training curriculum and self-supervised
learning objective in order to capture the structure and semantics of
tables.  Experiments across constrained and open-ended generation
settings show it is robust to different table sizes, performing
competitively and often better than state-of-the-art models. PixT3 is
also able to handle new domains with unseen tables, as evidenced by
our results on Logic2Text, a new dataset which we propose for
assessing the generalization capabilities of table-to-text generation
models.

Avenues for future research are many and varied. There are several
downstream tasks which stand to benefit from a pixel-based view of
textual information, including multilingual table-to-text generation,
and semantic parsing. We would also like to investigate additional
objectives and inductive biases that can better capture the structure
of tables and inter-cell dependencies.

\section{Limitations}
\label{sec:limitations}

While PixT3 shows promising results, its performance is affected by
the dimension of the input tables (for instance, 16\%~of the
Wikipedia tables in ToTTo remain too big for PixT3 to represent
effectively). It would be interesting to look into alternative ways of
preprocessing very large tables, e.g.,~by rendering them via multiple
images.  While our proposed intermediate training methodology
mitigates faithfulness errors, the model still struggles with
hallucinations, falling short of human-level performance. 

Finally, PixT3, as well as other comparison systems, have limited reasoning capabilities, e.g.,~they cannot infer information which is not explicitly stated in the table or make logical connections between concepts.  
PixT3's superior  performance in terms of faithfulness on Logic2Text (see Table~\ref{tab:human_eval})  is due to  generating simpler sentences rather than  superior reasoning skills. 
Thus, aside from new training objectives, a promising direction would be to combine the visual representations with an intermediate planning component that encourages the model to reason about the input while generating the output.




\section*{Acknowledgements}
We thank the meta-reviewer and anonymous reviewers for their constructive feedback. The authors also thank Ander Salaberria for his insightful comments on earlier versions of this work. We gratefully acknowledge the support of the UK Engineering and Physical Sciences Research Council (grant EP/L016427/1), the Basque Government (Research group funding IT-1805-22), MCIN/AEI/10.13039/501100011033 project AWARE (TED2021-131617B-I00), European Union NextGenerationEU/PRTR, and the LUMINOUS project (HORIZON-CL4-2023-HUMAN-01-21-101135724).



\bibliography{anthology,custom}

\appendix

\newpage
\section{Table Size Distribution in ToTTo}
\label{app:size_performance}

\begin{figure}[t]
\includegraphics[width=\linewidth]{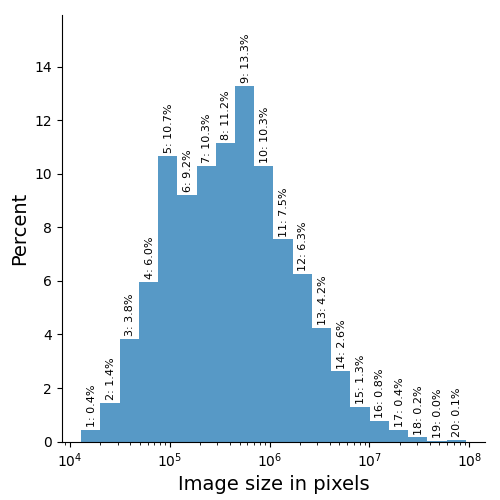}
\caption{Proportion of ToTTo examples (development set) per table size
  (shown in logarithmic scale).}
\label{fig:size_dist_log}
\end{figure}

We measure the size of a table by the total amount of pixels in its
corresponding rendered image. We then calculate the distribution of
each size, and group tables into 20~buckets accordingly.  Each bucket
covers a logarithmically increasing amount of table
sizes. Figure~\ref{fig:size_dist_log} shows the resulting buckets and
the proportion of ToTTo examples in each (development set). The
quality of descriptions generated within each group, are evaluated in
Section~\ref{sec:results}, see Figure~\ref{fig:size_performance}.

\section{Table-to-Text Generation Settings}
\label{sec:table-text-gener}

\begin{figure}[t]
\centering
\includegraphics[width=190pt]{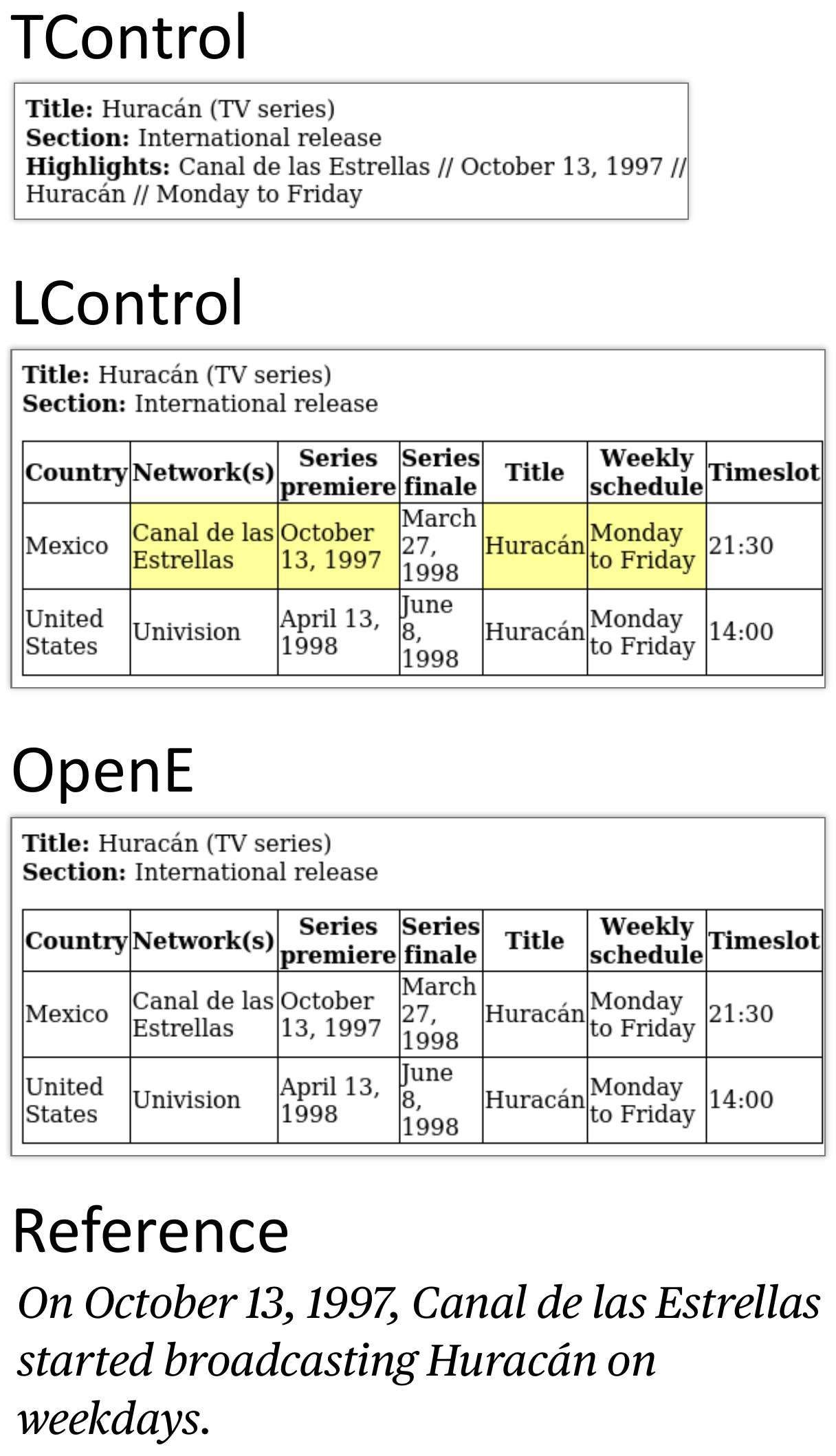}
\caption{PixT3 input image examples (and reference) in three
  generation settings: tightly controlled (TControl), loosely
  controlled (LControl), and open-ended (OpenE).}
\label{fig:examples}
\end{figure}

Figure~\ref{fig:examples} illustrates how the image input to PixT3
differs according to three generation settings: tightly controlled
(the model is given only highlighted cells, no table), loosely
controlled (the model is given the table and highlighted cells), and
open-ended (the model is given the table without any highlighting).

\section{Image Truncation and Down-scaling}
\label{sec:scale_truncation}
We explored the impact of down-scaling on model performance and its
tradeoff with truncation. We conducted a series of experiments wherein
PixT3 models were trained on versions of ToTTo with varying
down-scaling factor~$\gamma$: 0.87, 0.58, 0.39, 0.26, and 0.00. Note
that $\gamma$=0.00 corresponds to a setting where no truncation takes
place, only down-scaling. According to the results shown in
Table~\ref{fig:scale_trunc_exp}, it is best to combine truncation with
down-scaling, none of the extreme settings (no truncation vs too much
truncation) are beneficial. The optimal~$\gamma$ value is~0.39.


\begin{table}[t]
  \small
\begin{tabular}{l|*{5}{ccccc}} \toprule
\backslashbox{{Epoch}}{$\gamma$} & \makebox[2em]{{0.00}} & \makebox[2em]{{0.26}} & \makebox[2em]{{0.39}} & \makebox[2em]{{0.57}} & \makebox[2em]{{0.87}} \\ \hline
\hspace*{.2cm}16             & 28.71           & 29.13           & 29.47           & \textbf{29.58}  & 27.47           \\
\hspace*{.2cm}17             & 28.99           & 29.53           & \textbf{29.99}  & 29.70           & 27.69           \\
\hspace*{.2cm}18             & 29.67           & 30.04           & \textbf{30.55}  & 30.21           & 28.13           \\
\hspace*{.2cm}19             & 29.98           & 30.04           & \textbf{30.63}  & 30.54           & 28.33           \\
\hspace*{.2cm}20             & 29.83           & 30.21           & \textbf{30.68}  &
30.53           & 29.39           \\ \bottomrule
\end{tabular}
\caption{Evaluation results (BLUE) for PixT3 model in tightly
  controlled generation setting for different~$\gamma$ down-scaling
  factors. We show the Last five epochs on the  ToTTo  training
  set.}
\label{fig:scale_trunc_exp}
\end{table}

\section{Intermediate Training}
\label{app:ssl3}


\paragraph{Synthetic Dataset Generation} In this section we provide a
more detailed description regarding the generation of synthetic tables
for intermediate training. As our goal was to generate tables with a
structure similar to ToTTo, we first measured the probability
distribution of columns, rows, column spans and row spans for the
tables in the training set to avoid over-fitting and contamination.
We observed that the distribution of columns (up to 20~columns)
remained almost constant across tables, and did not affect the
probability distribution of rows. As a result, we aggregated row
numbers across columns and computed a single distribution for rows to
simplify our generation task, using discrete probability
distributions.  In order to limit the size of the generated tables we
cap the number of columns and rows to 20 and 75, respectively.  For
the synthetic text within the cells, we randomly generated digits in
the [1--5] range and character sequences from [A--Z, a--z] which gave
us a total of~776,520,240 permutations of possible unique cell values.

Overall, we generated 120K tables accompanied with target pseudo HTML
descriptions. The latter were on average 121~tokens long, with the
longest sequences containing 877~tokens. In experiments, we observed
that text size affects mainly the average count of tokens, whereas the
number of table columns and rows influences the length of the target
sequences. The sequences follow a hierarchical structure defined by the characters < and >. In the first hierarchical level, one container can be found for each highlighted cell in the table. Each container includes, in the following order, the highlighted cell, the cells in all related columns, and all cells in all related rows. This structure can represent multiple related columns and rows per highlighted cell, as well as multiple highlighted cells per table.

\begin{figure}[t]
\centering
\includegraphics[width=120pt]{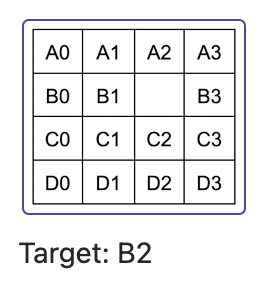}
\vspace{-.3cm}
\caption{Synthetically generated table with masked cell. Filled
  cell values denote position in the table.}
\label{fig:ssl3}
\end{figure}

\paragraph{Alternative Objectives}
We conducted a set of experiments to identify the best
self-supervised objective for our structure learning curriculum. In
addition to the objective presented in Section~\ref{sec:ssl1}, we also
experimented with a masking objective. Specifically, given a randomly
generated table, we filled each cell with text indicative of its
position in the table. We then masked random cells and the model was
trained to predict the missing cell values (see Figure~\ref{fig:ssl3}
for an example). We empirically observed that this objective led to
worse performance compared to PixT3, even though it resulted in
relatively fast training, since the table can be converted into a
sequence with a small number of tokens. We hypothesize that this
objective only weakly enforces table structure learning as the model
does not need to pay attention to all the cells in a column and row to
guess the missing value but simply rely on its closest neighbors. We
also experimented with a combination of the masking objective
discussed here and the structure learning objective described in
Section~\ref{sec:ssl1}. However, this model still lagged behind PixT3.

\begin{table}[t]
\centering
\resizebox{0.5\textwidth}{!}{
\begin{tabular}{@{}llcccc@{}}
& \textbf{}      & {Dev Set (All)} & {Test Set (Non)} & {Test Set (Over)} & {Test Set (All)} \\ \cmidrule(l){3-6} 
& {Model} & {BLEURT}        & {BLEURT}         & {BLEURT}          & {BLEURT}         \\ \midrule
& T5-base                            & 0.233                  & 0.106                   & 0.354                    & 0.230                   \\
& T5-3B                              & 0.228                  & 0.104                   & 0.344                    & 0.224                  \\
& Lattice                            & 0.226                  & 0.103                    & 0.348                     & 0.226                  \\
& CoNT                               & \textbf{0.240}         & 0.116                       & 0.364                        & 0.240                  \\
\raisebox{1em}[0pt]{\begin{sideways}TControl\end{sideways}} & PixT3                              & 0.178                  & 0.044                       & 0.312                        & 0.178                       \\ \midrule
& T5-base                            & $-$0.298                 & $-$0.395                  & $-$0.191                   & -0.293                  \\
& T5-3B                              & $-$0.309                 & $-$0.416                  & $-$0.194                   & -0.305                  \\
& Lattice                            & $-$0.287                 & $-$0.382                    & $-$0.195                     & -0.288                  \\
& CoNT                               & $-$0.293                 & $-$0.387                  & $-$0.190                   & -0.289                  \\
\raisebox{1em}[0pt]{\begin{sideways}LControl\end{sideways}} & PixT3                              & \textbf{0.169}         & \textbf{0.047}          & \textbf{0.287}           & \textbf{0.167}          \\ \midrule
& T5-base                            & $-$0.371                 & $-$0.458                  & $-$0.278                   & -0.368                  \\
& T5-3B                              & $-$0.385                 & $-$0.456                  & $-$0.301                   & -0.378                  \\
& Lattice                            & $-$0.377                 & $-$0.451                    & $-$0.302                     & -0.377                  \\
& CoNT                               & $-$0.370                 & $-$0.452                  & $-$0.281                   & -0.366                  \\
\raisebox{1em}[0pt]{\begin{sideways}OpenE\end{sideways}} & PixT3                              & \textbf{$-$0.332}        & \textbf{$-$0.414}         & \textbf{$-$0.258}          & \textbf{$-$0.336}        \\ \bottomrule
\end{tabular}
}
    \caption{BLEURT results on ToTTo for T5, PixT3, Lattice, and CoNT in three
      generation settings: tightly controlled (LControl), loosely
      controlled (LControl), and open-ended (OpenE). In the TControl
      setting, T5 results are taken from
      \citet{kale-rastogi-2020-text} and CoNT results from
      \citet{An2022}. This table complements results reported in
      Table~\ref{tab:totto_results}.}
    \label{tab:totto_bleurt_results}
\end{table}

\section{Additional Results and Examples}
\label{sec:addit-results-exampl}

\begin{figure*}[t]
\begin{tabular}{l@{}l}
  \includegraphics[width=120pt]{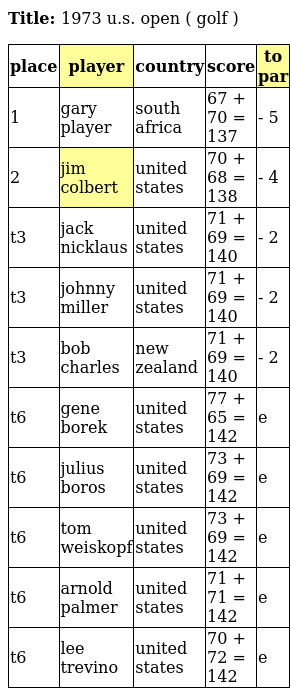} & 
\raisebox{5cm}[0pt]{\begin{minipage}{11.2cm}
\begin{itemize}
  \item \textbf{Reference:} Jim Colbert has the second best number of strokes to par.
  \item \textbf{CoNT (TControl):} Jim Colbert led the 1973 U.S. open (golf course) with a score of to par.
  \item \textbf{PixT3 (TControl):} Jim Colbert took part in the 1973 U.S. open (golf) tournament.
  \item \textbf{CoNT (LControl):} At the 1973 U.S. open (golf), Jim Colbert shot a record of 267 (+1) and finished four strokes ahead of runner-up Lee Janzen.
  \item \textbf{PixT3 (LControl):} Jim Colbert had a score of 142.
  \item \textbf{CoNT (OpenE):} Gary Player scored 137 (+5) and finished five strokes ahead of runner-up Jim Colbert.
  \item \textbf{PixT3 (OpenE):} Gary Player won the 1973 U.S. Open (golf) with a score of 137.
\end{itemize}
\end{minipage}}
\end{tabular}
\caption{Logic2Text table and model output in three generation
  settings: tightly controlled (TControl), loosely controlled
  (LControl), and open-ended (OpenE).}
\label{fig:tab_example_l2t}
\end{figure*}

In addition to BLEU and PARENT reported in
Tables~\ref{tab:totto_results} and~\ref{tab:logic2text_results}, we also present results with BLEURT in
Table~\ref{tab:totto_bleurt_results} and Table~\ref{tab:logic2text_results_bleurt}. We further show example output
on the Logic2Text dataset (zero-shot setting) in
Figure~\ref{fig:tab_example_l2t}. In the TControl setting, CoNT
struggles to produce a coherent sentence, while PixT3 generates a
faithful but not very informative one. This is not surprising as the
models receive nothing but the title and highlighted cells, making it
extremely difficult to generate the target sentence. In LControl, both
models have access to the entire table; however, they still produce a
false statement, most likely a consequence of the zero-shot nature of
our generation task. Finally, in the less constrained OpenE setting,
PixT3 generates a coherent and faithful sentence. While CoNT also
produces a fluent sentence, it incurs a faithfulness error when
mentioning "(+5)" instead of "(-5)". This is likely due to the
performance degradation this model experiences when provided with the
full table.


\begin{table}[t]
\begin{small}
\begin{center}
  \begin{tabular}{@{}llc@{}} \toprule
& {Model} &  {BLEURT} \\ \midrule
& LLaVA                  & $-$1.230          \\
& T5-base               & $-$1.086          \\
& T5-3B                  & $-$1.079 \\
& Lattice                  & \textbf{$-$1.060}          \\
& CoNT                 & $-$1.103          \\
\raisebox{1.2em}[0pt]{\begin{sideways}TControl\end{sideways}} & PixT3          & $-$1.104          \\ \midrule
& LLaVA                  & $-$1.189          \\
& T5-base                  & $-$1.147          \\
& T5-3B                   & $-$1.167          \\
& Lattice                  & $-$1.147          \\
& CoNT                  & $-$1.159          \\
\raisebox{1.2em}[0pt]{\begin{sideways}LControl\end{sideways}} & PixT3          & \textbf{$-$1.073} \\ \midrule
& LLaVA                  & \textbf{$-$1.184} \\
& T5-base               & $-$1.237          \\
& T5-3B                 & $-$1.196          \\
& Lattice               & $-$1.231          \\
& CoNT           & $-$1.231          \\
\raisebox{1.6em}[0pt]{\begin{sideways}OpenE\end{sideways}} & PixT3           & $-$1.213          \\ \bottomrule
\end{tabular}
\end{center}
\end{small}
    \caption{Automatic evaluation results on Logic2Text in three
      generation settings: tightly controlled (LControl), loosely
      controlled (LControl), and open-ended (OpenE). All models
      (except LLAVA) were fine-tuned on ToTTo
       and tested on the Logic2Text. This table complements results reported in
      Table~\ref{tab:logic2text_results}.}
    \label{tab:logic2text_results_bleurt}
\end{table}

\section{LLaVA promts}
\label{sec:llava_promts}
As mentioned in Section~\ref{sec:experimental-setup}, our zero-shot
experiments involved comparisons against LLaVA-1.5
\cite{liu2023visual}, a large pretrained multimodal model (13B
parameters). We devised the following prompts for each generation
setting:

\paragraph{TControl}
"Here are some descriptions based on other highlights of other tables
'chilawathurai had the 2nd lowest population density among main towns
in the mannar district .', 'zhou mi only played in one bwf super
series masters finals tournament .', 'tobey maguire appeared in vanity
fair later than mike piazza in 2003 .'. Now write a short description
based on the following highlighted cells extracted form a table."

\paragraph{LControl} "Here are some descriptions based on the
highlights of other tables not present in the input: 'chilawathurai
had the 2nd lowest population density among main towns in the mannar
district .', 'zhou mi only played in one bwf super series masters
finals tournament .', 'tobey maguire appeared in vanity fair later
than mike piazza in 2003 .'. Now write a short description based on
the highlighted cells in this table following the same style as the
example descriptions."

\paragraph{OpenE} "Here are some descriptions from other tables not
present in the input: 'chilawathurai had the 2nd lowest population
density among main towns in the mannar district .', 'zhou mi only
played in one bwf super series masters finals tournament .', 'tobey
maguire appeared in vanity fair later than mike piazza in 2003 .'. Now
write a short description stating something from this table following
the same style as the example descriptions."

\section{Human Evaluation Guidelines}
\label{app:form}
We provide the full set of instructions presented to crowdworkers for the human evaluation
study. Our participants were native English speakers from the United Kingdom and the United States of America, with a 50/50 equal 
gender split between male and female. 

\textsf{
Thank you for taking part in our experiment! You will be presented with a table and a computer-generated description of its content. Your task is to determine whether each description is "True" or "False" based on the information provided in the table and/or its title and subtitle (you will see examples later-on). No expert knowledge is required to perform this task. You should evaluate the descriptions given the information presented in the table, without taking any other information into account (e.g., based on your own knowledge or the web).}

\textsf{Here are some guidelines to help you with your evaluation:}

\textsf{\textbf{Acronyms}: tables often have acronyms which the descriptions might spell out.  For example, if the table mentions "TD" and the description correctly spells it out as "touch down," you should not consider this "False" (although the description might be false for other reasons).}

\textsf{\textbf{Implicit information}: the description might mention information that can be inferred but is not explicitly spelled-out in the table.  For example, it could mention "steam engines" when the table lists theirs names without explicitly talking about steam engines. In this case, the description should not be considered "False".}

\textsf{- You should evaluate each description independently.}

\textsf{- If the description does not make sense and is impossible to evaluate (usually when summarizing very large tables), you should consider it as "False".}

We suggest starting by reading the description and then referring to the table to verify if it aligns with its claims.

\textsf{This data elicitation study is performed by researchers at [REDACTED].  If you have any questions, feel free to contact [REDACTED]. Participation in this research is voluntary. You have the right to withdraw from the experiment at any time. The collected data will be used for research purposes only. We will not collect any personal information.  Your responses  will be linked  to your anonymous Prolific ID for the exclusive purpose of conducting our experiment.}


\section{PixT3 Fine-tuning Hyper-parameters}
\label{app:hyperparams}

PixT3 models across all three settings (TControl, LControl, OpenE) were fine-tuned using the same hyper-parameters. To prevent over-fitting, we employed early stopping based on the BLEU score computed on the validation set every 250 steps. Table \ref{tab:hyper} enumerates the specific hyper-parameter values used in PixT3, with all remaining parameters set to  the default values defined in Pix2Struct \cite{Lee2023}.

\begin{table}[]
\centering
\begin{tabular}{p{4cm}p{2cm}}
\toprule
\textbf{Hyperparameter} & \textbf{Value} \\
\midrule
Optimizer &  AdamW\\
Learning rate  & 0.0001 \\
Warm-up steps & 1000 \\
Max. input patches & 2048 \\
Shuffle train data & False \\
Epochs  & 30 \\
Train batch size  & 8\\
Gradient accum. steps  & 32\\
Mixed precision & fp16 \\
Evaluation batch size  & 32 \\
Eval freq. steps & 250 \\
Inf. beam search & 8 beams\\
\bottomrule
\end{tabular}
\caption{Hyperparameters used in PixT3.}
\label{tab:hyper}
\end{table}

\end{document}